\documentclass{article}
\usepackage{spconf,amsmath,epsfig}
\usepackage{xspace}
\usepackage{multirow}
\usepackage{cite}

\makeatletter
\DeclareRobustCommand\onedot{\futurelet\@let@token\@onedot}
\def\@onedot{\ifx\@let@token.\else.\null\fi\xspace}

\def\ie{\emph{i.e}\onedot}

\def\etal{\emph{et al}\onedot}
\makeatother

\begin{document}\sloppy

\def\x{{\mathbf x}}
\def\L{{\cal L}}

\title{Skeleton Boxes: Solving skeleton based action detection with a single deep convolutional neural network}
%
\name{Bo Li$^{1*}$, Huahui Chen$^{1}$, Yucheng Chen$^{1}$, Yuchao Dai$^{2}$, Mingyi He$^{1*}$
      \thanks{$^{*}$ libo.npu@gmail.com, myhe@nwpu.edu.cn}
\address{$^{1}$School of Electronics and Information, Northwestern Polytechnical University, China \\
         $^{2}$Research School of Engineering, Australian National University, Australia \\
         }
}

\maketitle

\begin{abstract}
Action recognition from well-segmented 3D skeleton video has been intensively studied. However, due to the difficulty in representing the 3D skeleton video and the lack of training data, action detection from streaming 3D skeleton video still lags far behind its recognition counterpart and image-based object detection. In this paper, we propose a novel approach for this problem, which leverages both effective skeleton video encoding and deep regression based object detection from images. Our framework consists of two parts: skeleton-based video image mapping, which encodes a skeleton video to a color image in a temporal preserving way, and an end-to-end trainable fast skeleton action detector (Skeleton Boxes) based on image detection. Experimental results on the latest and largest PKU-MMD benchmark dataset demonstrate that our method outperforms the state-of-the-art methods with a large margin. We believe our idea would inspire and benefit future research in this important area.
\end{abstract}
\begin{keywords}
skeleton, detection, CNN, end-to-end
\end{keywords}
\section{Introduction}
\label{sec:intro}
As an intrinsic high level representation, 3D skeleton is valuable and comprehensive for summarizing a series of human dynamics in the video, and thus benefits the more general action analysis\cite{Liu2017PKU}. Recently, 3D skeleton activity analysis has drawn great attentions~\cite{Du2015Hierarchical, Veeriah2015Differential, Zhu2016Co, Shahroudy2016NTU, Liu2016Spatio, Song2016An,Du2016Representation}. However, due to the difficulty in representing the 3D skeleton video and the lack of training data, action detection from streaming 3D skeleton video still lags far behind its recognition counterpart and image-based object detection.

Different from the skeleton activity recognition from well-segmented video clip, skeleton action detection is more difficult, due to the need of recognition and location simultaneously. Although it is of great importance, there are very few works specially designed for it. And it is still an open and critical question that what is a proper framework for this problem.

Recently, deep learning based methods have achieved great success in the computer vision field, especially those high-level recognition problems. While, how to design an efficient action detection system that leverages the neural network for the skeleton-based data is not well studied.

\begin{figure*}[!htb]
\centering
\includegraphics[width=0.95\linewidth]{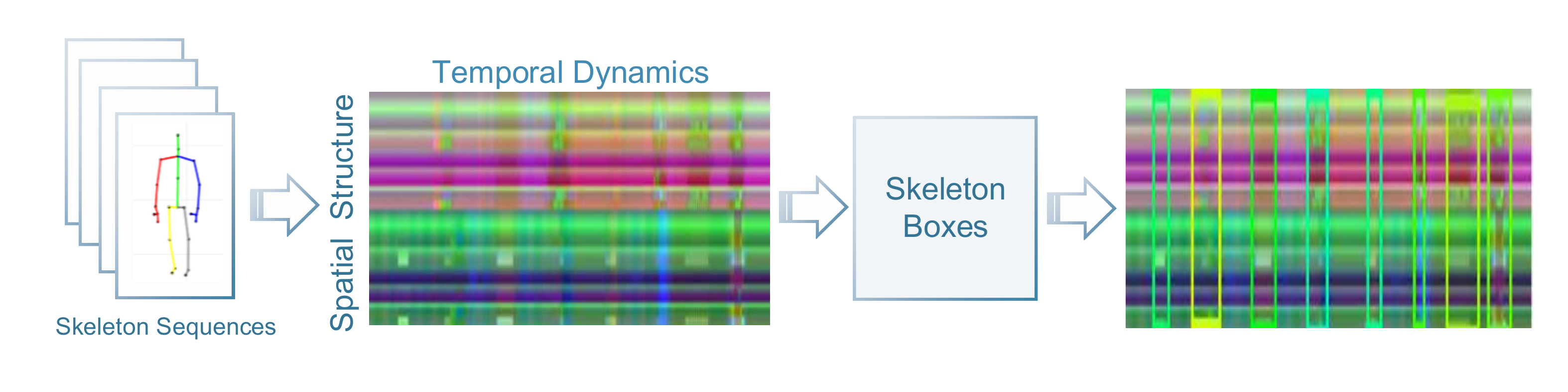}
\caption{Flowchart of the proposed method.}
\label{fig:flowchart}
\end{figure*}

In this paper, we propose a novel end-to-end approach for skeleton-based action detection problem. Our work partially inspired by the recent success of general image object detection with CNN~\cite{liu2016SSD} and skeleton video image mapping method in \cite{Yong2015Skeleton}. A conceptual illustration of our framework is presented in Fig \ref{fig:flowchart}.

Here we would like to mention some very related work, and declare the main difference from them. In the work of~\cite{Yong2015Skeleton}, a skeleton video image mapping strategy is proposed for skeleton based recognition problem. This work inspires us, while we modified the mapping strategy. And more importantly, their work's focus is on the recognition problem, we extent it to the detection problem. Very recently, Li \etal~\cite{li2016online} proposed an online Human Action Detection method which utilized a joint Classification-Regression Recurrent Neural Network. Their method could account for the online detection problem. Compared with \cite{li2016online}, we adopt totally different framework and our offline performance is better than theirs. Our work is also related to the SSD~\cite{liu2016SSD} which is an effective general image object detection method, and widely utilized in other image detection work~\cite{poirson2016fast, liao2016textboxes}. Our work is built on this framework, and we find this framework could also work well on skeleton mapping image, although it is totally different from the nature image. More importantly, we  have made some specially modification to make it more adaptive to the skeleton based data. Thus we call our method ``Skeleton Boxes", which is a variety of SSD and specially designed for skeleton data.

In conclusion, our contributions could be summarized as following:

\begin{itemize}
  \item Based on the skeleton data image mapping, a novel end-to-end CNN based detection approach for skeleton-based data is proposed. Our method is totally end-to-end and avoids the typical sliding window or proposal design and thus ensures high computational efficiency.
  \item We achieve the state-of-the-art result on the largest PKU-MMD benchmark and outperform other methods by a large margin.

\end{itemize}

\section{Method}

Our method consists of two parts. (1) the translation-scale invariant image mapping. (2) SkeletonBoxes. A conceptual illustration of our framework is presented in Fig.\ref{fig:flowchart}.

\subsection{Translation-scale invariant image mapping}
Similar to the work in \cite{Yong2015Skeleton}, we divide all human skeleton joints in each frame into five main parts according to human physical structure, \ie two arms, two legs and a trunk. To preserve the local motion characteristics, joints in each part are concatenated as a vector by their physical connections. Then the five parts are concatenated as the representation of each frame. To map the 3D skeleton video to an image, a natural and direct way is to represent the three coordinate components $(x,y,z)$ of each joint as the corresponding three components $(R,G,B)$ of each pixel in an image. Specially, each row of the action image is defined as $R_i = [x_{i1},x_{i2},...,x_{iN}]$, $G_i = [y_{i1},y_{i2},...,y_{iN}]$, $B_i = [z_{i1},z_{i2},...,z_{iN}]$, where $i$ denotes the joint index and $N$ indicates the number of frames in a sequence. By concatenating all joints together, we obtain the resultant action image representation of the original skeleton video.

Due to the coordinate difference in 3D skeleton and image, proper normalization is needed. Du \etal \cite{Yong2015Skeleton} proposed to quantify the float matrix to discrete image representation with respect to the entire training dataset. Specifically, given the joint coordinate $c^{jk}$ ($x,y,z$), the corresponding pixel value $p^{jk}$ is defined as
\begin{equation}
\label{softmax}
p^{jk} = floor\left(255 * \frac{c^{jk}-c_{min}}{c_{max}-c_{min}}\right),
\end{equation}
where $c_{max}$ and $c_{min}$ are the maximum and minimum of all joint coordinates in the training set respectively, $jk$ represent the $k$-th channel ($x,y,z$) of the $j$-th skeleton video sequence. $floor$ is the rounding down function. 


\begin{figure}[htb]
\centering
\includegraphics[width=0.85\linewidth]{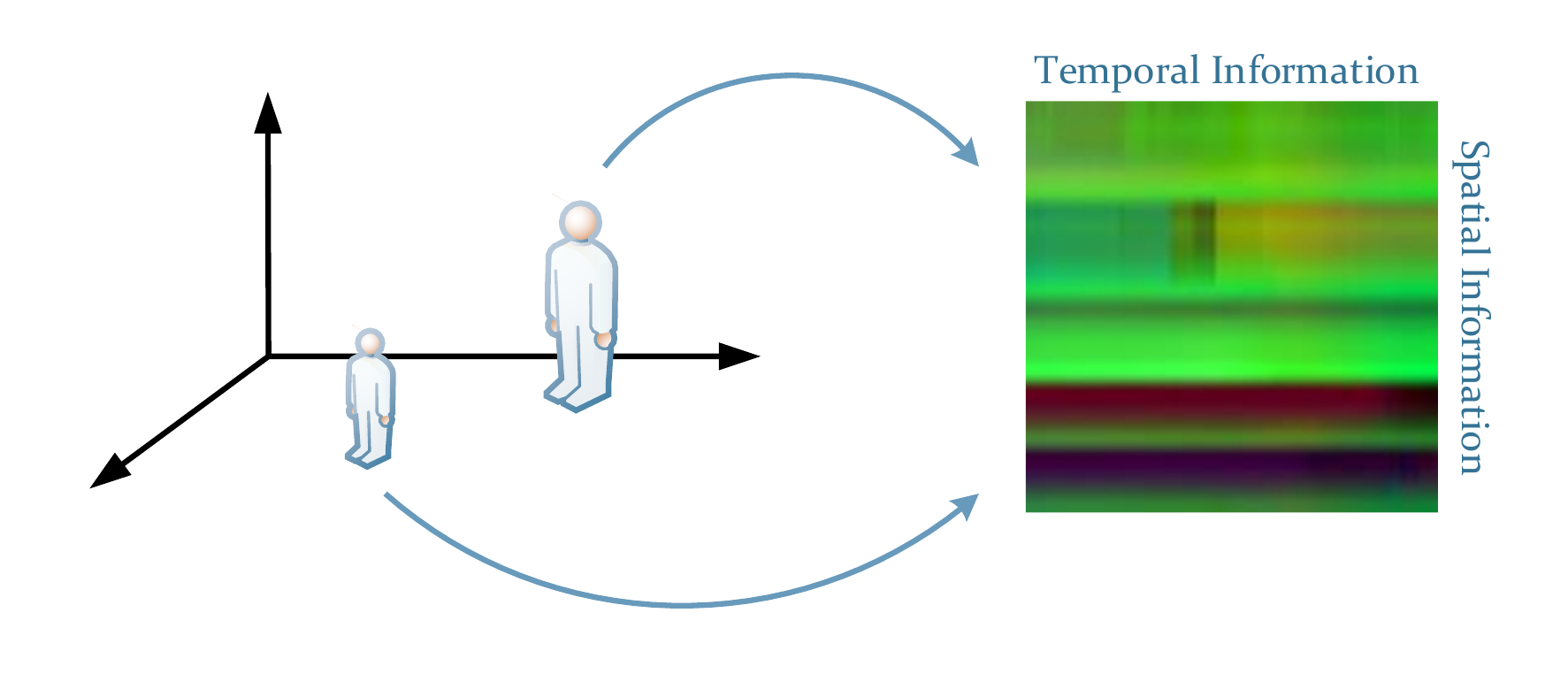}
\caption{Illustration of the translation-scale invariant image mapping.}
\label{fig:mapping}
\end{figure}

Here, we propose a simple and yet effective translation-scale invariant image mapping,
\begin{equation}
\label{eq:our_mapping}
p^{jk} = floor\left(255 * \frac{c^{jk}-c^{jk}_{min}}{\max\limits_{k}(c^{jk}_{max}-c^{jk}_{min})}\right),
\end{equation}
where $c^{jk}_{max}$ and $c^{jk}_{min}$ are the maximum and minimum coordinate value of the $k$-th channel ($x,y,z$) of the $j$-th skeleton video sequence.

Compared with Du \etal \cite{Yong2015Skeleton}, our image mapping method owns the following properties:  1) \textbf{Translation invariant:} Our image mapping transforms the 3D coordinates with respect to the minimum coordinate in each sequence rather than the entire training dataset, thus the translation in 3D does not affect the action video as illustrated in Fig.~\ref{fig:mapping}. 2) \textbf{Scale invariant:} Similarly, by normalizing the 3D skeleton coordinates with respect to the 3D variation along the three axis in each sequence, the scale change has also been eliminated. 3) \textbf{Dataset independent:} As the minimum and maximum coordinates are extracted from each sequence independently, the normalization is thus independent to each specific skeleton video sequence. Lastly, our normalization is isometric to each coordinate $x,y,z$, thus the relative scale in different axis has been well preserved.

\subsection{Skeleton Boxes}

\subsubsection{Architecture}
\begin{figure}[htb]
\centering
\includegraphics[width=0.85\linewidth]{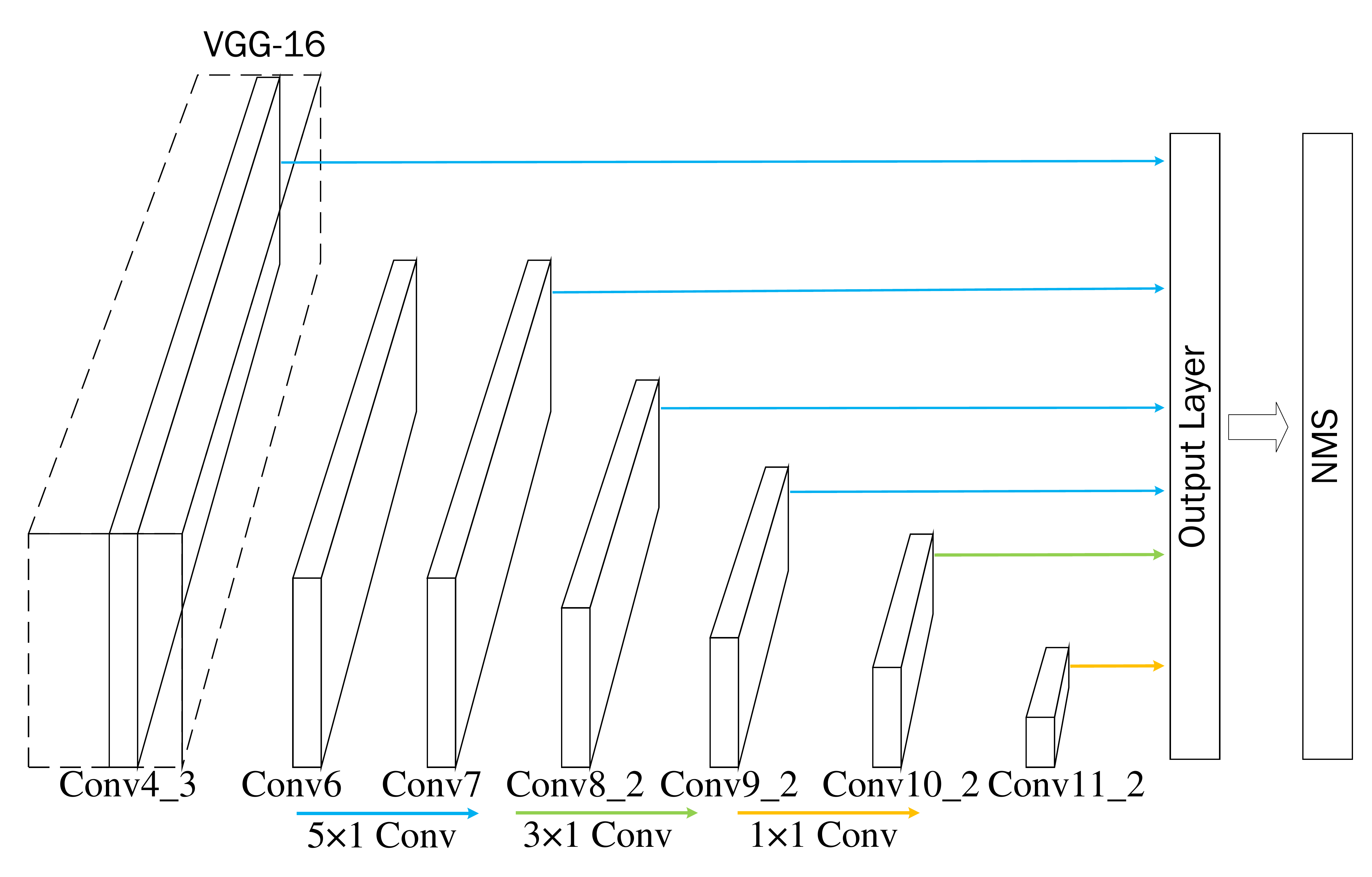}
\caption{Illustration of the architecture of our CNN network. NMS denote non-maximum supression. }
\label{fig:net_archt}
\end{figure}

The architecture of our Skeleton Boxes is illustrated in Fig.\ref{fig:net_archt}. Our Skeleton Boxes is built on the SSD framework~\cite{liu2016SSD}. We make some modifications to adjust the Skeleton-based detection problem. Although these contributions may seem small independently, they improve the performance. In this paper, we choose the popular VGGNet as our backbone networks for two reasons. Firstly, the VGGNet is much shallower than the ResNet, thus it is much computational cheap than the ResNet. Secondly, our experiments also show that the performance of ResNet is worse than VGGNet in this task. As the work of~\cite{liu2016SSD}, we keep the layers from \emph{conv1\_1} through \emph{conv4\_3}. The last two fully-connected layers of VGG-16 are converted into convolutional layers. They are followed by a few extra convolutional and pooling layers, namely \emph{conv8} to \emph{conv11}. An illustration of our network architecture is presented in Fig.\ref{fig:net_archt}.

\subsubsection{Detection kernel}

In order to detect the skeleton action clip of multiple scale or length, multiple output layers are inserted after the last and some intermediate convolutional layers. In this work, we adopt irregular $5\times1$ convolutional filters instead of the standard $3\times3$ ones. This specially designed convolution filter would better fit skeleton action clips with large aspect ratios. In addition, it also tends to avoid noisy signals that a square-shaped receptive field would bring in. The outputs of the network are aggregated and undergo a non-maximum suppression (NMS) process in order to get the final detection results.

\subsubsection{Aspect ratios for skeleton boxes}
In the training phase, ground-truth word boxes are
matched to default boxes according to box overlap, following the matching scheme in~\cite{liu2016SSD}. Each map location is associated with multiple default boxes of different aspect ratios. They effectively divide skeleton video clips by their scales and aspect ratios. Within the SSD framework, how to design the optimal tiling is an open question as well. And the design of default boxes is highly task-specific.

Different from general objects, skeleton boxes tend to have large
aspect ratios. Therefore, we include ``high'' default boxes that have large aspect ratios. Specifically, we define 6 aspect ratios for default boxes, including $\frac{1}{7},\frac{1}{5},\frac{1}{3},\frac{1}{2},1,2,3,5,7$.

\subsubsection{Learning}
We adopt the same loss function as \cite{liu2016SSD}. Let
$x$ be the match indication matrix, $c$ be the confidence, $l$ be the predicted location, and $g$ be the ground-truth location. Specifically, for the $i$-th default box and the $j$-th ground truth, $x_{ij} = 1$ means matching while $x_{ij} = 0$ otherwise. The loss function is defined as:
\begin{equation}
\label{softmax}
L(x,c,l,g) = \frac{1}{N} (L_{conf}(x,c) + \alpha L_{loc}(x,l,g))
\end{equation}
where $N$ is the number of default boxes that match ground-truth boxes, and $\alpha$ is set to 1. We adopt the smooth $L_1$ loss for $L_{loc}$ and a multi-class softmax loss for $L_{conf}$.

Hard negative mining and data augmentation is also conducted in our training phase. With the hard negative mining, we keep the ratio between the negative and positive is at most $3:1$. The data augmentation method utilized in our training phase is mainly randomly patch sampling.

\section{Experiments}
The proposed method is evaluated on the PKU-MMD dataset. To our best knowledge, this is the largest skeleton-based data action detection dataset. It contains almost 1076 long video sequences in 51 action categories, performed by 66 subjects in three camera views. For fair comparison and evaluation, the same dataset partition setting as that in \cite{Liu2017PKU} was used.

\textbf{Cross-View Evaluation:} For cross-view evaluation, the videos sequences from the middle and right Kinect devices are chosen for training set and the left is for testing set. For this evaluation, the training and testing sets have 717 and 359 video samples, respectively.

\textbf{Cross-Subject Evaluation:}  In cross-subject evaluation, we split the subjects into training and testing groups which consists of 57 and 9 subjects respectively.

Our experiments are conducted on the CNN toolbox:caffe~\cite{Jia2014Caffe}. Our network is trained by SGD. We choose momentum of 0.9 and weight decay of 0.0005. The initialization learning rate is set to 0.000004, and divided by 10 when the loss is not decrease and repeated for 3 times. The batch-size is set to 4.

\subsection{Error metrics}
For quantitative evaluation, we report errors obtained with the following error metrics, which have also been used in~\cite{Liu2017PKU}.

\begin{equation}
\frac{|I\cap I^{\ast}|}{|I\cup I^{\ast}|} > \theta
\end{equation}
where $I\cap I^{\ast}$ denotes the intersection of the predicted and ground truth intervals and $I\cup I^{\ast}$ denotes their union. So, with $\theta$, the $p(\theta)$ and $r(\theta)$ can be calculated.



\textbf{Mean Average Precision (mAP):}  With several parts of
retrieval set $Q$, each part $q_{j} \in Q$ proposes $m_{j}$ action occurrences ${\{d_{1} , . . . d_{{m}_{j}} \}}$ and $r_{jk}$ is the recall result of ranked $k$ retrieval results, then mAP is formulated by
\begin{equation}
\label{softmax}
\textnormal{mAP}(\theta) = \frac{1}{|Q|} \sum_{j=1}^{|Q|} \frac{1}{m_{j}} \sum_{k=1}^{m_{j}} p_{interp}(r_{jk},\theta)
\end{equation}

In this paper, we adopt the mean average precision of different actions.
%

\subsection{Experimental results}

Table \ref{tab:results} lists the performance of the proposed method and those state-of-the-art results. It is obvious that our method outperforms other methods by a large margin.

\begin{table}[htb]
\center
\caption{mAP comparison of results among several approaches on PKU-MMD dataset. The correspondent results are quoted directly from \cite{Liu2017PKU}}

\begin{tabular}{  r || c | c || c | c }
\hline
{Method} & \multicolumn{2}{c||}{Cross-view} & \multicolumn{2}{c}{ Cross-subject}\\
\hline
{$\theta$} & {0.1} & {0.5} & {0.1} & {0.5} \\
\hline
{SVM}  &{0.240} &{0.181} &{0.036} &{0.021}\\
\hline
{BLSTM}  &{0.545} &{0.479} &{0.159} &{0.130}\\
\hline
{STA-LSTM} &{0.476} &{0.444} &{0.155} &{0.131}\\
\hline
{JCRRNN}  &{0.699} &{0.533}  &{0.452} &{0.325}\\
\hline
{Proposed Method}  &\bf{0.945} &\bf{0.942}  &\bf{0.613} &\bf{0.548}\\
\hline
\end{tabular}
\label{tab:results}
\end{table}

\section{Conclusion}
Based on the skeleton data image mapping, a novel end-to-end CNN based detection approach for skeleton-based data is proposed. Our method is totally end-to-end and avoids the typical sliding window or proposal design and thus ensures high computational efficiency. Experiments on the challenging PKU-MMD dataset show that our method outperforms the state-of-the-art methods by a large margin.

\vspace{3mm}
\textbf{Acknowledgement}
{
This work was supported in part by Natural Science Foundation of China grants (61420106007, 61671387) and Australian Research Council grants (DE140100180).
}
\small
\bibliographystyle{IEEEbib}
\bibliography{ref}

\end{document}